# SPIKE UP PRIME INTEREST IN SCIENCE AND TECHNOLOGY THROUGH CONSTRUCTIONIST GAMES

P. Petrovič, F. Agarshev

*Comenius University Bratislava (SLOVAKIA)*

## Abstract

Robotics sets have been successfully used in elementary and secondary schools in conformance with the 'learning through play' philosophy fostered by LEGO Education, while utilizing the Constructionism didactic approach. Learners discover and acquire knowledge through first-hand tangible experiences, building their own representations in a constructivist learning process. Usual pedagogical goals of the activities include introduction to the principles of control, mechanics, programming, and robotics [1]. They are organized as hands-on learning situations with teamwork cooperation of learners, project-based learning, sharing and presentations of the learners group experiences. Arriving from this tradition, we focus on a slightly different scenarios: employing the robotics sets and the named approaches when learning Physics, Mathematics, Art, Science, and other subjects. In carefully designed projects, learners build interactive models that demonstrate concepts, principles, and phenomena, perform experiments, and modify them in elaboration phases with the aim to connect, create associations and links to the actual underlying theoretical curriculum. In this way, they are collecting practical experiences which are prerequisite to successful learning process. Based on feedback from children, we continue upon two previous sets of activities that focused on Physics and Mathematics, this time with projects built around games. Learners play various games with physical artifacts in the real-world - with the models they build. They acquire skills while playing the games, analyze them, and learn about the underlying principles. They modify the game rules, strategies, create extensions, and interact with each other in an entertaining and engaging settings. This time we have designed the activities together with the children, students of applied robotics seminar, and an undergraduate student of Applied Informatics. In this way, we have immediately received feedback and verified their applicability and feasibility.

## 1 INTRODUCTION

We have been organizing educational robotics competitions for more than 20 years, and these can be divided into two main categories: *prepare and compete* and *come and play with us*. In the first category, the participation in the contest is a long-term project lasting several weeks up to a year, or more, where participants work hard in their clubs, homes and schools to design, build, program, test and debug their robots. At the end they come to a tournament, which is a kind of a social event, a celebration concluding that long period of work. However, it sometimes leads to a lot of stress as a small problem can and often leads to a complete failure of the robots resulting in frustration and disappointment from several months of work "wasted". The second category is a different style – participants work throughout the year on any other creative projects, and for the contest they take their robotic sets, come to a tournament, get a challenging task and work independently from any external help or support for a couple of hours. They do their best, some of them receive awards, but having learned what they could from the experience, they can forget about it the other day and continue their usual activities. Both styles have their pros and cons but taken from the point of view of maximizing the learning benefit for 10+ participants in a robot club, the first type of challenges has certain consequences: it allocates most of the resources of the whole group, and the leader, focusing on a single outcome for a long period of time. Clearly, the involvement of the individual members would be uneven and very difficult to manage in such a way that all participants would follow a personally optimal learning trajectory. Certainly, it is a very valuable one-time experience occasionally, but we feel it is not so suitable as a long-term strategy in a robot club. Eleven years ago, we have therefore invented a novel robot competition that is a mixture of the two styles: Robot League [2,3]. RL is an on-line competition with about 20 different challenging tasks per semester published on-line. Teams work remotely for the period of about one month to solve them and publish their solutions to the on-line platforms: description, pictures, programs, video. Then – in compliance with the principles of the Constructionism – the solutions are publicly shared so that the teams – after working hard on that challenge for several weeks – can see alternative methods of tackling the challenge from other teams. The robot club leader can thus divide the group into multiple subgroups, each working on a suitable challenge at their level of experience, and enjoying both the benefits of longer-term project work, and avoiding the stressful fragile situations at a final tournament. For example, a challenge in the first round in 2014 – before the age of gyro sensors being part of robot sets – was to build a "robot skier" that is placed on a tilted surface, it should turn uphill, climb up, and "ski" down. The task for the teams was to invent some way of detecting the "uphill" direction only using light, distance, and touch sensors.

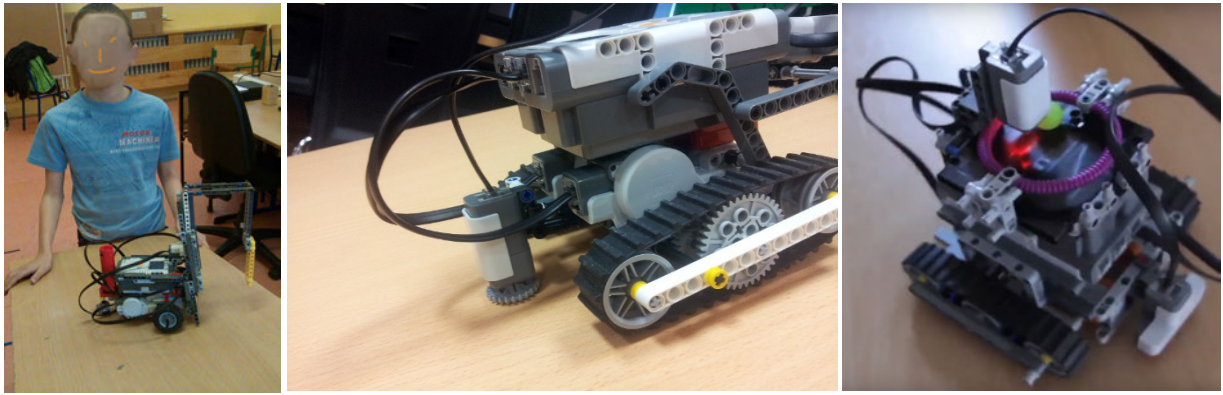

*Figure 1. Different solutions to a "skier" task: measuring distance from a freely hanging brick (left), flat U-shaped caterpillar robot leaning to one side with a touch sensor (middle), and a rolling ball on top of the robot detected by a light sensor (right). From solutions of teams Tobias, Amazing Team, and Gamčabot.*

Some solutions used a freely hanging LEGO brick harnessing the gravity force which would either extend outside or inside of the robot depending on the robot orientation on the slope. They detected that using a light or distance sensor. Another group would use a kind of inverted pendulum bouncing to sides and pressing touch sensors. Yet, another group built a caterpillar robot with flat U-shape caterpillars causing the robot to lean forward or backwards depending on its orientation on the slope – and again detecting that with the touch sensor. The cleverest solution involved a circular space containing a freely rolling ball and a light sensor capable of detecting the ball and positioned above one end of that space – the location the ball reaches exactly when the robot is oriented uphill, see figure 1. In fact, this group of students invented and built a new kind of sensor! All the solutions required and improved the participants abilities of thinking out of the box, creativity, and building and programming skills.

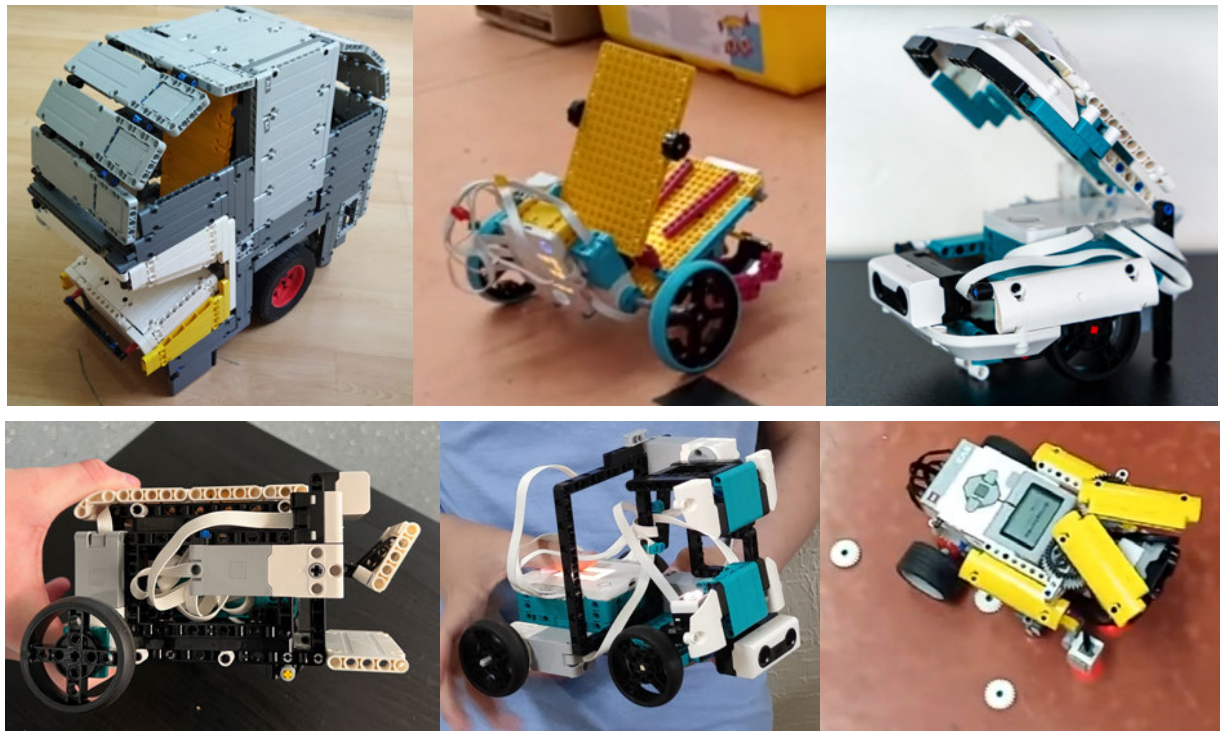

*Figure 2. Several solutions to a Pac-Man task from teams Rubik's bots, Šachisti, Haloworld, Boomers, Dubáci, and Legolas.*

Until now, there have been about 140 similar different tasks with hundreds of children benefiting from this program – run with very low budget and by volunteers with occasional sponsor contributions that are used for prizes. In the fourth round of 2022, the participants were to build Pac-Nan robots and program a real Pac-Man game that takes place in a real world as contrasted to traditional on-screen computer game, see figure 2. We have received feedback from the participants that this was the most interesting task in the season. Not only did they learn a lot, but they also built something that they had some good fun with. This was an important moment that spawned the efforts described in this paper.

During the past two years, we have been working on finding ways of using the robotic sets in educational settings in interdisciplinary scenarios. We designed set of projects for demonstrating and/or discovering concepts in Physics and Mathematics [4,5,6]. In the follow-up of these projects, we started to design LEGO educational games. Games that can be built, played, and tinkered with – modified by changing rules, adding new features, solving little or more advanced tasks. This approach has multiple benefits: 1) it provides the students and teachers with interesting examples of projects at moderate-level complexity – for inspiration and learning; 2) it provides extra motivation, because of the fun aspect of playing the games with the built models – even whole class tournaments can be arranged; 3) it trains the students to work with legacy code (read, modify), which are very important and useful skills; 4) it stimulates the creativity of students through open-ended tasks associated with the models; 5) it also shows how to document their models: each model is modelled in 3D modelling software stud.io, step-by-step building instructions are available, with pictures, videos, and further materials. In the remaining sections of this paper, we will briefly review related literature, describe the projects with have created with the Scratch-like Word Blocks programming language with children in a robot club, mention the projects made by our students of Applied Informatics at robot seminar, and focus on Python projects developed as part of one of the author's final thesis, touch on games children has submitted to a special round of our Robot League competition, where they designed their own games, discuss the results and impact before our concluding remarks.

## 2   RELATED WORK

Learning by playing is a very old concept that has been visited in many periods of human history. For instance, Plato seemed to suggest that intellectual play in some form, as demonstrated in the dialectical banter of Socrates, could provide a stimulus to understanding [7]. In the 17$^{th}$ century, John Amos Comenius, an education reformer and today known as a Father of Modern Education suggested that teaching methods should appeal to the whole person. His work, rediscovered in late 19th century [8], is accepted worldwide today. The whole philosophy of LEGO Education is based on learning through play [9]. To stress the significance of play in learning, specific terms have been coined, such as serious games or Papert's hard fun. Playful situations have been successfully utilized in method of Milan Hejný to teach Mathematics [10]. In the recent decades, a substantial amount of literature, dedicated conferences and journals focus on using games in the context of education, sometimes called game-based learning, while application of game playing into all kinds of processes including education is labeled gamification. For example [11] study the adoption of serious games in formal education settings.

While a lot of scholars consider the sole working with educational robotics sets already as a play or a kind of a game, specific works where realistic real-world games with the aspect of tangible interactive human-robot interaction are not so common, they may focus on some peculiar aspect and usually create only one isolated and simple game. Some works may contain a lot of pedagogical research bloat yet missing our target of truly creative interactive games. For example, in [12], participants of the experiment interact with a LEGO robot with an arm to manipulate ball and learn about loops. In [13], authors designed an educational game utilizing LEGO robots. Android phones programmed in MIT App Inventor run the game "NXTTriviaRacing" involving a combination of quiz games (trivia) with simple vehicle racing. Authors notice that although children are used to high-quality digital games on phones and computers, if the game interacts with robotics constructions that the children themselves have made, this changes their attitude. An interesting study of tangible physical games interconnected with digital devices using RFID technology is in [14]. Authors designed three prototypes a) Block-Magic, an educational game based on well-known Logic Blocks containing 64 brightly colored and durable plastic blocks, b) Walden PECS Communicator, a platform based on Picture Exchange Communication System (PECS) a worldwide methodology to enhance communication skills in autistic persons, and c) WandBot, a learning environment that combines LEGO NXT robots, RFID-technology and serious games for scientific dissemination in science centres. For instance, in a) each augmented magic block had an integrated/attached passive RFID sensor for wireless identification of each single block. This allowed them to construct a trainer, capable of seeking the specific learning curve of each child and Adaptive Tutor System providing instruction that were tailored on individual learners. Inspiring studies can be found in HCI field. For instance, in [15] authors study applications of robotic educational games in the community of visually impaired children. They designed a game for spatial navigation and mobility that is inclusive of both visually impaired and sighted children. Their guidelines emphasize multisensory feedback, hands-on creation, and narration as a means for modulating pace and stimulation; all of which were highlighted as potential barriers to inclusive interactions.

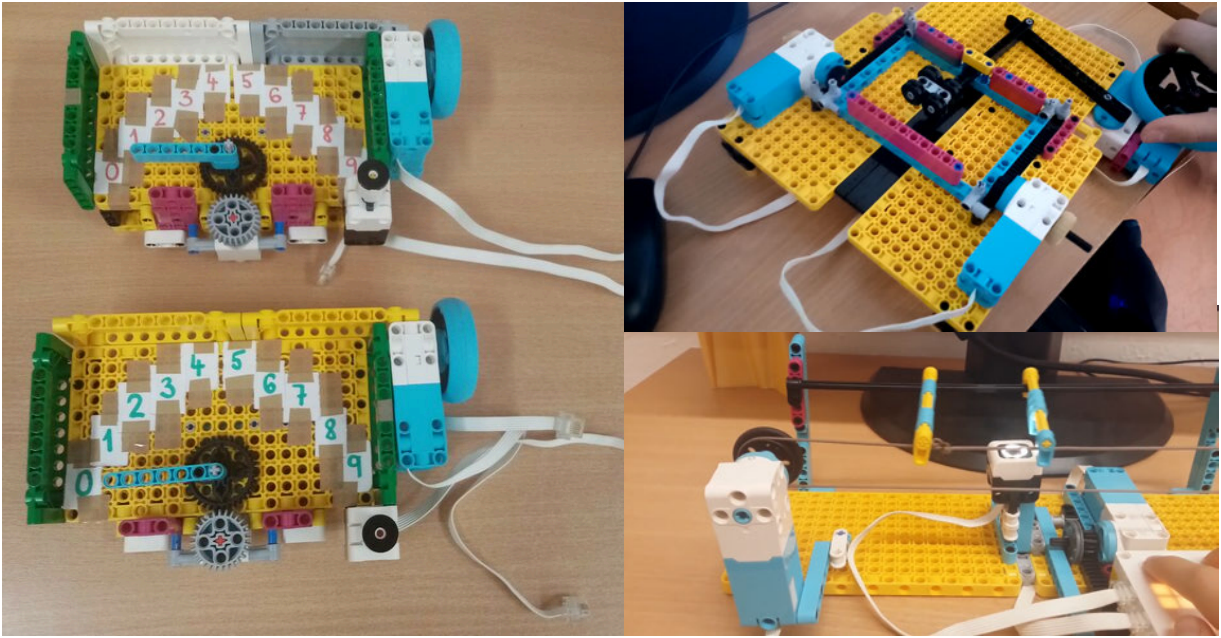

*Figure 3. Example projects from the set of projects with Word blocks programming language.*

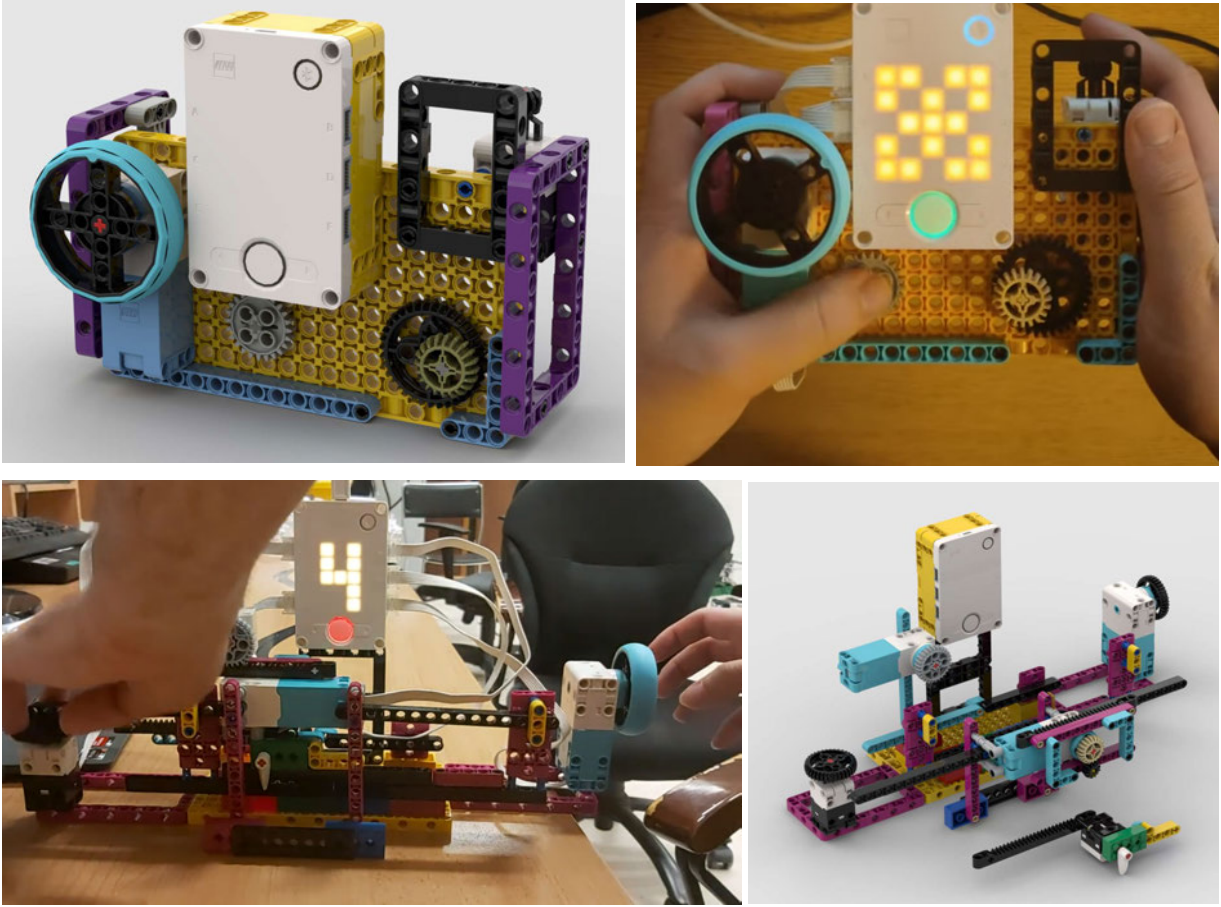

*Figure 4. Student projects – Battle Ships top, and Color Select bottom.*

On the official support pages, LEGO Education provides series of didactic materials that include building instructions, student supplementary materials, teacher guidelines organized according to 5E instruction method [16]. We will adopt this style also for most of the projects designed here, in addition, we provide 3D virtual models designed with Stud.io software so that the users can study and modify the details of the models themselves. Other great resources of teaching materials always been those of Tufts Center for Engineering Education and Outreach [17], and CMU Robot Academy [18].

## 3  CREATIVE PROJECTS WITH WORD BLOCKS

Immediately after the Pac-Man round, we started to think about some games to design with the 10-16 years old children in our robot club. Figure 3 left shows a game for two players: both players select a number from 0-99 (the interval is variable). The hand pointing to a digit is controlled by turning a wheel on the side, and a digit is confirmed by pressing a touch sensor. The player who selects higher number wins, unless it is more than twice the other player's number. The players continue for several rounds, until they decide to close the game. The LEGO hub calculates and reports the moves and scores of both players throughout. Other examples of such games are shown on figure 3 right. These are ideas and realizations completely by children – the frame representing a road is moving left and right (top) and the objects attached to a string are moved left and right, in both cases the task is to avoid collisions by controlling the movement either with a dial or buttons on the LEGO hub.

Next semester, we asked two 2$^{nd}$ year students of Applied Informatics at our Faculty who were taking Practical Robotics Seminar course as their final project in the course to create games with Spike Prime sets – with the building instructions, Stud.io models, programs, and ideas for further improvements. Their projects are shown in figure 4 - a strategic Battleships game and a color memory game Color Select. We have later verified the projects with the participants of our robot club.

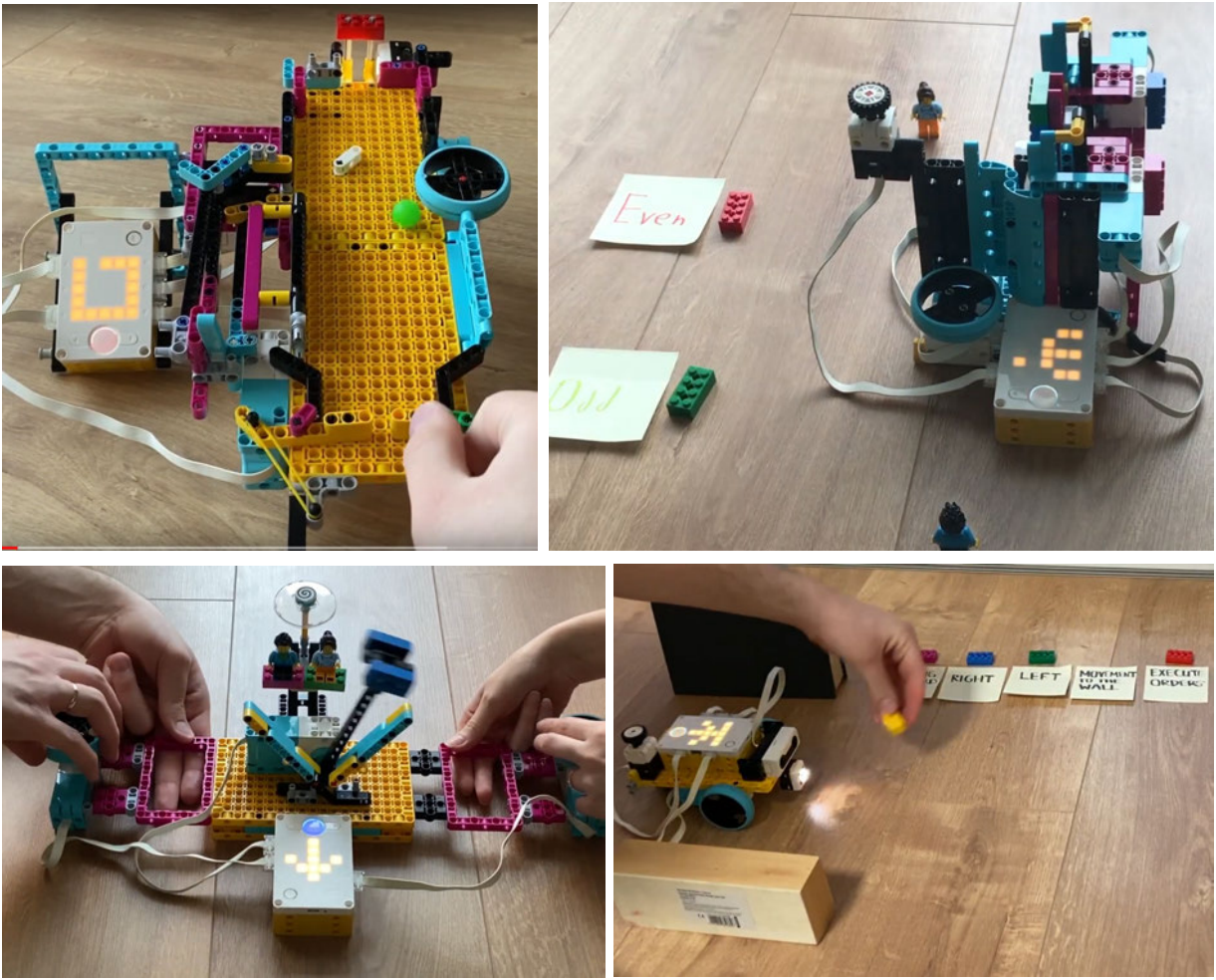

*Figure 5. Example projects from the set of projects with Python programming language.*

## 4 CREATIVE PROJECTS WITH PYTHON

In a more systematic approach, one of the authors selected a final thesis project [19] to design a set of such game projects, but this time, the models were programmed in Python programming language. Our motivation was to contribute to the selection of Spike Prime projects with Python, which is not a very rich one, and to make the Spike Prime technology available for secondary school students, who are missing a suitable platform after the MINDSTORMS EV3 support ended. Each of these projects can be built from a single robotic set, and it contains: 1) a 3D model designed in Stud.io, 2) building instructions to construct the model, 3) a Python program for the basic version of the game; 4) pictures and videos of the model; 5) tasks for the students on modifying and improving the program or the model; 6) instructions for the teacher (lesson plan) organized in 5E instructional model methodology, and 7) supporting material for the student – typically instructions. The requirement for design of these games was that the playful interaction must occur in the real world with physical artefacts moving, and the user physically interacting with them. All of them thus should contain tangible interfaces and perform real actions. We believe this is a vital aspect, children acquiring experiences in the real world, creating connections with different areas of their mind improving on their creativity and the potential for associative reasoning in conformance with Constructionism approach. Figure 4 contains four examples of Spike Prime Python game projects: 1) a pinball game utilizing sensors and motors in interaction of a popular game, 2) a guessing game with betting, 3) probably the most popular model of a pong game, which is entertaining to play since it requires certain skill and as the players play it more, they are constantly improving. We used this model in public exhibits of our university open day, and Maker Faire in Vienna. And 4) a little color-programming maze robot to improve the abstract thinking and imagination of the player.

Other games developed in this batch were the popular game Tic-Tac-Toe and a storage Safe. The first utilizing search algorithms such as Minimax, the second working with data structures representing the secret key. All projects were designed with the goal for the students to improve their Python programming skills, discover and become familiar with new features of the language. All these projects have been made public at the same location as the previous two sets of projects [6]. We have verified these projects with students in a secondary school, see below.

## 5 ROBOT LEAGUE GAME TASK 2023

To further investigate the idea of games constructed of LEGO robotics set, we have used of the 5 rounds in this year's Robot League and specified the following task [20] for the participants: "*Build a model using your robot set. Program it so that the result will be some game that can be played either by one or two players. Something must be happening in the real world, not only on the hub's display, so you will use the motors and sensors. It can move, a player can hold it in his or her hands, or it can be placed on a table. It should be possible to gain some scores, which will be reported. The game should be fun, not too complex, and not too easy. The most important criterion (in addition to playability) is originality – i.e. it is not easy to find something similar, the idea is your own*." For the more advanced teams, we also asked them to create a virtual model in Stud.io and the building instructions. The teams could freely choose to build a game, or to solve another more standard task, the maximum score from the two tasks count, so some teams solve both tasks. Some teams used the Spike Prime sets, others used the EV3. Out of 23 participating teams only 2 choose not to make a game (and one of them had already built a game that was used as an example before this round). This proves the apriori strong motivational factor of the game models idea. Some teams were limited in the time available as their participation in the competition is organized by their informatics teacher and they work directly in the classes. Other teams that work in robot clubs or at home had much more time. Some teams constructed more than one game. Some games were quite simple, and they would require some more work to serve as models for others yet had some inspiring idea. Some were more advanced and interesting. Let's study the ideas of the participants in detail: In 5 cases they have built fast-reaction games plus 2 table-top rolling fast car track with collision avoidance, 4 were about practicing precision or speed of action, 2 pinballs, 1 hockey and 1 football game, 3 remotely controlled robot-cars (chasing, slalom, sumo), 2x rock-paper-scissors, and 1 color-memory game. Almost all satisfied the requirement of task specification of using motors, sensors, and programs. Some teams recorded and submitted videos of playing a tournament, clearly, this task was a successful one with a lot of fun and learning going on. The number of teams participating in this third round did not change significantly (1st round: 20 teams, 2nd round: 24 teams), and while we usually see a decline in the number of teams through the season due to competition with other robot contests in that period, no decline has occurred this time.

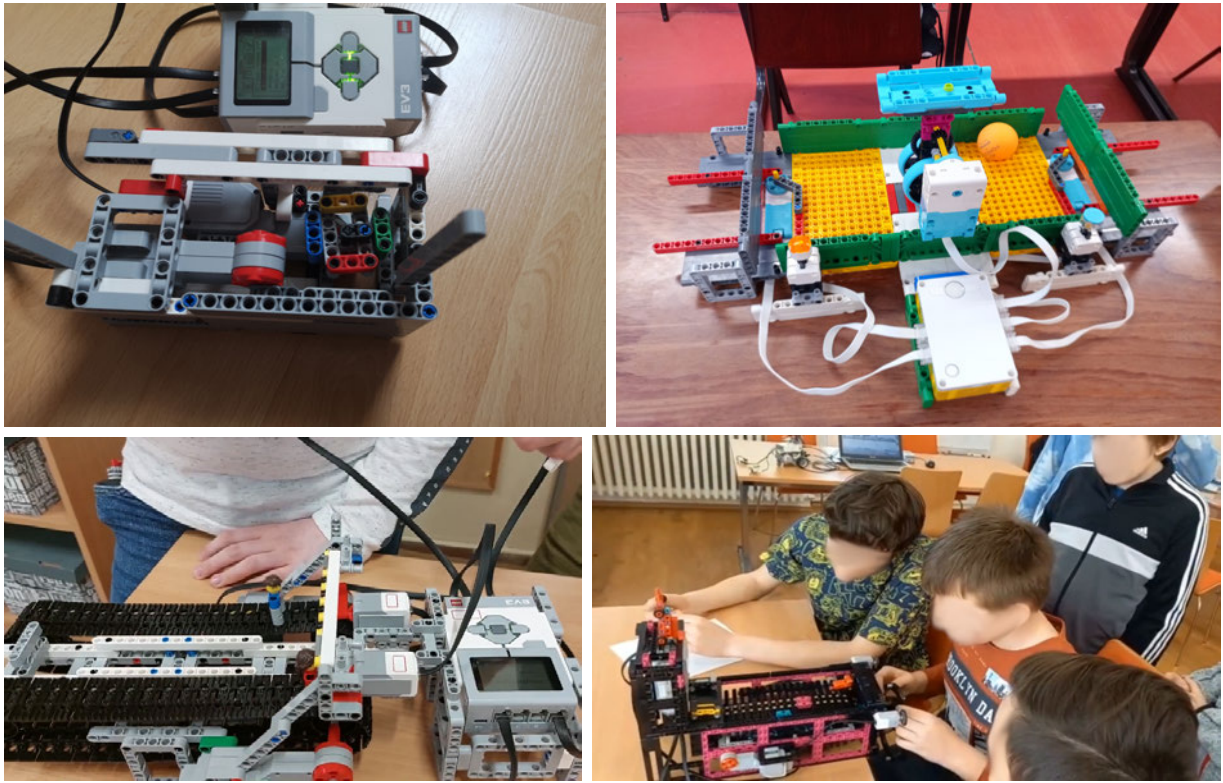

*Figure 6. Example projects from 3<sup>rd</sup> round of Robot League 2023 – Task Game.*

Figure 5 contains selected solutions from teams HomBoti, Legolas, SvitBoti, and Tomato Robot 3: the color memory game, hockey, fast-action game (competing LEGO figures), and table-top avoidance car track, which the kids used to run a local tournament with their constructed game.

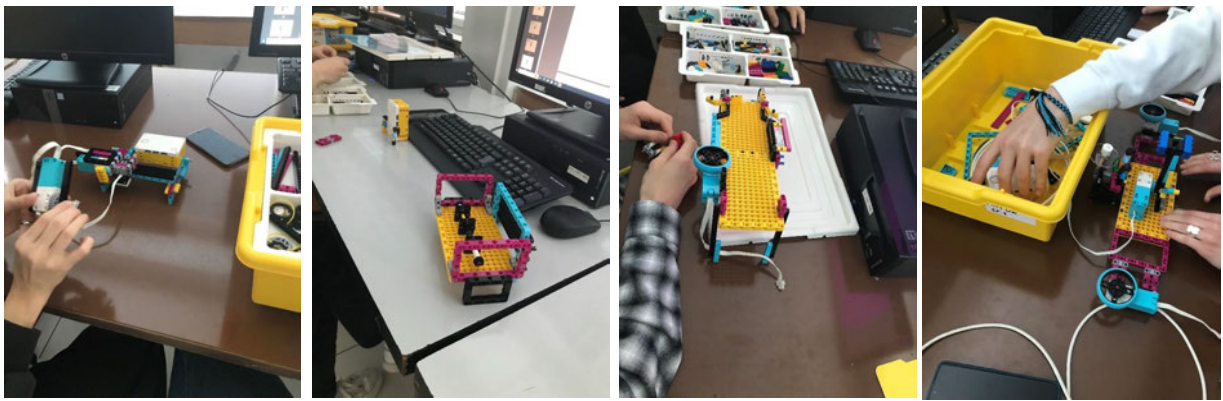

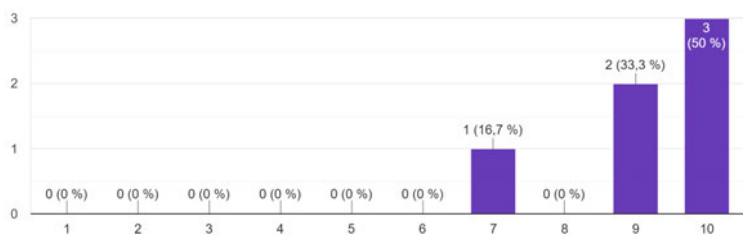

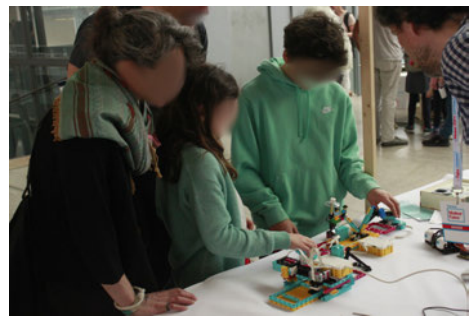

*Figure 7. Evaluating the developed games with students (top), their feedback (bottom left), presentation at Maker Faire Vienna (bottom right).*

# 6  RESULTS

In addition to small experiments in our robot club, and positive experiences at Maker Faire event in Vienna in early May 2023, we have evaluated the Python projects from the final thesis in a class in the Gymnázium ul. Ladislava Sáru in Bratislava. 12 students aged 15-17, 2/3 of being girls, worked in pairs, see figure 7. Each pair building and playing with a different game they selected on their own. We could use a double 45-minute slots. For most of them the LEGO building and programming was a new concept, and thus the time required was a bit longer than with experienced subjects and we used some of the time for an introduction and explanation how to operate the hub and download a program. The responses shown in the chart suggest the students enjoyed the lesson. The qualitative feedback they provided was also positive, mostly reporting they liked to learn how different parts work and fit together, they liked the building part, enjoyed how they discovered the principle of operation of the machine, and that it was fun and interesting. They also gave a few hints that we could use to improve the instructions and found a calibration procedure required in one model as less interesting. Since they worked in a regular computer lab, the room was also not perfect for arranging the boxes of the robotic sets around their working table. In the forthcoming evaluations, we will focus on the how the models improve the programming skills of the students when working on the model programming challenges.

# 7  CONCLUSIONS

The main theme of our work is studying interdisciplinary situations for using educational robotics sets in teaching. As a follow-up of previous two studies with projects targeted on supporting teaching in Physics and Mathematics, this work focuses on projects that are real-world playable games. Some of the projects were designed by us, some by our students, some by participants in elementary-school robot club, and some by participants in the 3rd round of this year's Robot League competition that we have been organizing for 11 years. Some of the designed projects focused on improving algorithmic thinking and programming skills in Python language and aim to be useful contributions to the Python community of LEGO Spike Prime robotic set users – since at the time of development of these projects the LEGO Education Spike Prime application transitioned from version 2 (with support towards end of 2023) to incompatible version 3, the projects contain source-code for both versions. Each game serves as a tangible interactive constructionist model. It is controlled by its own modifiable rules and its project contains programming or building tasks for the elaboration phase of the 5E instructional model, as well as virtual 3D model, detailed building instructions, teacher and student supporting material. During the early evaluations of the designed models, we have received positive feedback, and plan on adding more projects, and finding more interdisciplinary situations where robot sets could be used – already in other rounds of this year's Robot League we have included some projects with an overlap with Biology and Ethics. We also see various connections to the field of human-robot interaction, where we already plan performing more controlled studies with subjects to measure qualitative and quantitative impact.

## ACKNOWLEDGEMENTS

The project was partially supported by Horizon Europe project TERAIS, GA 101079338.